# Tabular PDF Information Extraction with Local LLMs and Layout-Aware Parsing: A Reliability Evaluation

**Muhammad Anis Al Hilmi*[1], Neelansh Khare[2], Noel Framil Iglesias[3], Kurnia Adi Cahyanto[4], Azhar Al Afghani[1], Musfi Yuliadi[1]**

[1]Informatics, Faculty of Engineering, Universitas Swadaya Gunung Jati, Cirebon, Indonesia
[2]Computer Science, University of California, Irvine, USA
[3]Computer Engineering, UNIR, La Rioja, Spain
[4] Doctoral Program of Information Systems, Universitas Diponegoro, Semarang, Indonesia

Email: [1*]muhammadanisalhilmi.dosen@ugj.ac.id

**Abstract**

Extracting structured information from academic PDF documents is non-trivial: a single page typically combines free-text metadata with tabular regions, exhibits cross-program variation, and is susceptible to Unicode encoding artifacts that interfere with downstream parsing. This study evaluates the reliability of information extraction approaches for tabular PDF documents, using academic course registration documents (Kartu Rencana Studi, or KRS) from Indonesian higher education as a case study. Three strategies are compared: LLM-only, Hybrid Deterministic–LLM (regex + LLM), and a Camelot-based pipeline with LLM fallback. Experiments were conducted on 140 documents for the LLM-based test and 860 documents for the Camelot-based pipeline evaluation, covering four study programs with varying data in tables and metadata. Three 12–14B LLM models (Gemma 3, Phi 4, and Qwen 2.5) were run locally using Ollama and a consumer-grade CPU without a GPU. Evaluations used exact match (EM) and Levenshtein similarity (LS) metrics with a threshold of 0.7. Although not applicable to all models, the results show that the hybrid approach can improve efficiency compared to LLM-only, especially for deterministic metadata. The Camelot-based pipeline with LLM fallback produced the best combination of accuracy (EM and LS up to 0.99–1.00) and computational efficiency (less than 1 second per PDF in most cases). The Qwen 2.5:14b model demonstrated the most consistent performance across all scenarios. These findings confirm that integrating deterministic and LLM-based methods is a reliable and efficient strategy for information extraction from tabular text-based PDF documents in computationally constrained environments.

## 1. INTRODUCTION

PDF documents containing embedded tabular data are pervasive across many domains [1]—invoices, financial statements, scientific reports, and academic records [2], among others [3], [4]. Reliably extracting structured information from such documents remains a long-standing problem in document processing [5]: the table region typically requires different handling from surrounding free-text metadata [6], and PDFs frequently introduce encoding artifacts (most notably Unicode ligatures arising from font substitution) that disturb naive text extraction. Classical deterministic methods, such as regular expressions, rule-based parsers, and layout-aware libraries, handle well-formed cases efficiently but tend to be brittle in the face of structural variation [7]. The recent availability of large language models (LLMs) that can be executed locally on consumer hardware has opened a complementary direction [8], raising the question of how reliable each approach is—individually and in combination—on real tabular PDF documents, particularly under constrained hardware where local execution offers cost and privacy advantages over cloud-based alternatives.

To investigate this question concretely, this study uses academic course registration documents (Kartu Rencana Studi, or KRS) issued by Indonesian higher education institutions as a case study. In Indonesia, students complete a KRS at the start of each semester to select their courses, with each course associated with a credit unit (SKS) and a designated lecturer [9]; program rules range from fixed block systems [10] to free elective schemes [11], and the resulting KRS is typically saved as a PDF [12]. KRS PDFs are a convenient test bed for the problem outlined above: they are single-page, text-based rather than scanned, and combine an unstructured metadata region at the top (student identity, study program, semester, and academic year) with a course table in the body. They also exhibit non-trivial variation across study programs in the number of courses, in lecturer name formats, and in academic degree abbreviations. They are subject to the same Unicode ligature artifacts [13] as other text-based PDFs (for example, the letter pair "fi" rendered as the single glyph "ﬁ"). This combination of regularity and controlled variation makes KRS a compact yet genuinely challenging dataset for evaluating tabular PDF information extraction pipelines.

### 1.1. Related Works

In general, extracting information from PDFs can be done in several ways. Some PDFs are text-based, meaning they can be copied from the PDF, while others are derived from images/scans/OCR results [14]. Furthermore, there are structured, semi-structured, and unstructured formats [1]. Academic documents generally fall into the structured category. Approaches based on OCR, rule-based parsing (regex), and machine learning [15] have been widely used in the legal [1], [7], [16], health [17], [18], log parsing [19], [20], [21],[22],[23], fiscal document [24], and identity card domains. However, academic documents, such as study plans (KRS), exhibit variations in data across study programs, tables, and encoding issues (e.g., Unicode ligatures) that can interfere with the extraction process. As LLM (Learning and Learning) develops, prompt-based parsing approaches have become an attractive alternative due to their ability to understand context.

Research over the past five years has shown a trend in the use of LLM [25] and OCR for document extraction across various domains. Some studies have utilised OCR to read driver's licenses, legal documents, and clinical data [17], while others have highlighted the efficiency of regex for stable elements [18]. Studies on Chinese [26] character parsing have emphasised the importance of Unicode normalisation and encoding.

Recent research has also addressed few-shot prompting for table parsing [27], free-OCR to reduce reliance on commercial engines, and evaluation of LLM in structured extraction tasks. However, most studies focus on general or medical domains; studies of Indonesian academic documents, particularly KRS, remain very limited. Based on the literature, there are several research gaps: (1) limited systematic evaluation of tabular PDF information extraction pipelines that compare deterministic, LLM-only, and hybrid approaches on a single dataset; (2) limited comparative evaluation between LLM-only and hybrid deterministic–LLM approaches for documents that combine free-text metadata with tabular regions and exhibit Unicode encoding artifacts; (3) limited evaluation of local LLM execution on consumer CPU devices for documents containing personal data; and (4) lack of analysis of throughput and stability of results. This study aims to fill these gaps.

## 2. METHOD

If Observations of the target/object of the problem revealed the following situation in the PDF of the student's KRS (Course Plan) file. The PDF consists of a single page with a header at the top containing the university logo and identity. The text data in this KRS file is PDF-based (not image- or OCR-based), meaning it can be copied, blocked, and selected directly. The top text, which contains study program information, student personal data, semester, and academic year, is not in a table layout. The middle text, including the list of courses and lecturers' names, is contained within a table, clearly separated by line breaks. The data varies, including numbers, letters, symbols, and some letters that appear as Unicode (due to the font), for example, "fi" appears as "ﬁ" (ligature). This presents a challenge. Categorising the data, three types will be extracted: metadata/student personal information, course list, and lecturer list.

The experiment was conducted using a Python script, the Fitz library for PDF text extraction, and Ollama as a local LLM runtime. The models used were Gemma 3:12b, Phi-4:14b, and Qwen 2.5:14b. All experiments were run on a consumer-grade CPU laptop without a GPU. Laptop software and specifications: Advan Workplus. Ryzen 5 6000H Series CPU. 16GB RAM. 1TB SSD. Windows 11 Home 64-bit OS. Ollama 0.12.3. Python 3.12.4. Camelot 1.0.9.

The dataset was prepared with three variations: four different study programs, variations in the LLM model, and variations in modes (deterministic with regex, Camelot library, LLM, and a combination). Data variations across study programs include: semester, number of courses, variation in lecturer academic degrees, variation in student and lecturer name length, and multiple rows of data in a single column. The number of PDFs per study program was 35, totalling 140 for the testing process. A total of 860 PDF files were prepared, accompanied by ground-truth (GT) JSON files generated by parsing a cloud-based enterprise model (note: for privacy reasons, the service does not store the data) and manually cross-checked.

For processing, a variety of models were selected, tailored to specific CPU levels. Models were selected from major players: Microsoft, Alibaba, and Google. Llama from Meta was not selected because it was briefly tested but computationally intensive, making it unlikely to run on the laptop used. The models were Gemma3:12b, Phi4:14b, and Qwen2.5:14b. There were also various modes. Regex was used as a deterministic method (for fast computation, at the cost of adjusting the number of rules) to reduce the LLM's computational load. Camelot, a library that claims to understand tabular and text data (not scanned/OCR), was also used to complement the regex and combination processing.

## 2.1. Pipeline Scheme

Figure 1 is system pipeline. The testing pipeline consists of: (1) PDF text extraction; (2) biodata extraction using regex (hybrid mode); (3) course and lecturer table extraction using LLM; (4) Unicode/ligature normalisation; (5) post-process JSON validation; and (6) ground truth evaluation.

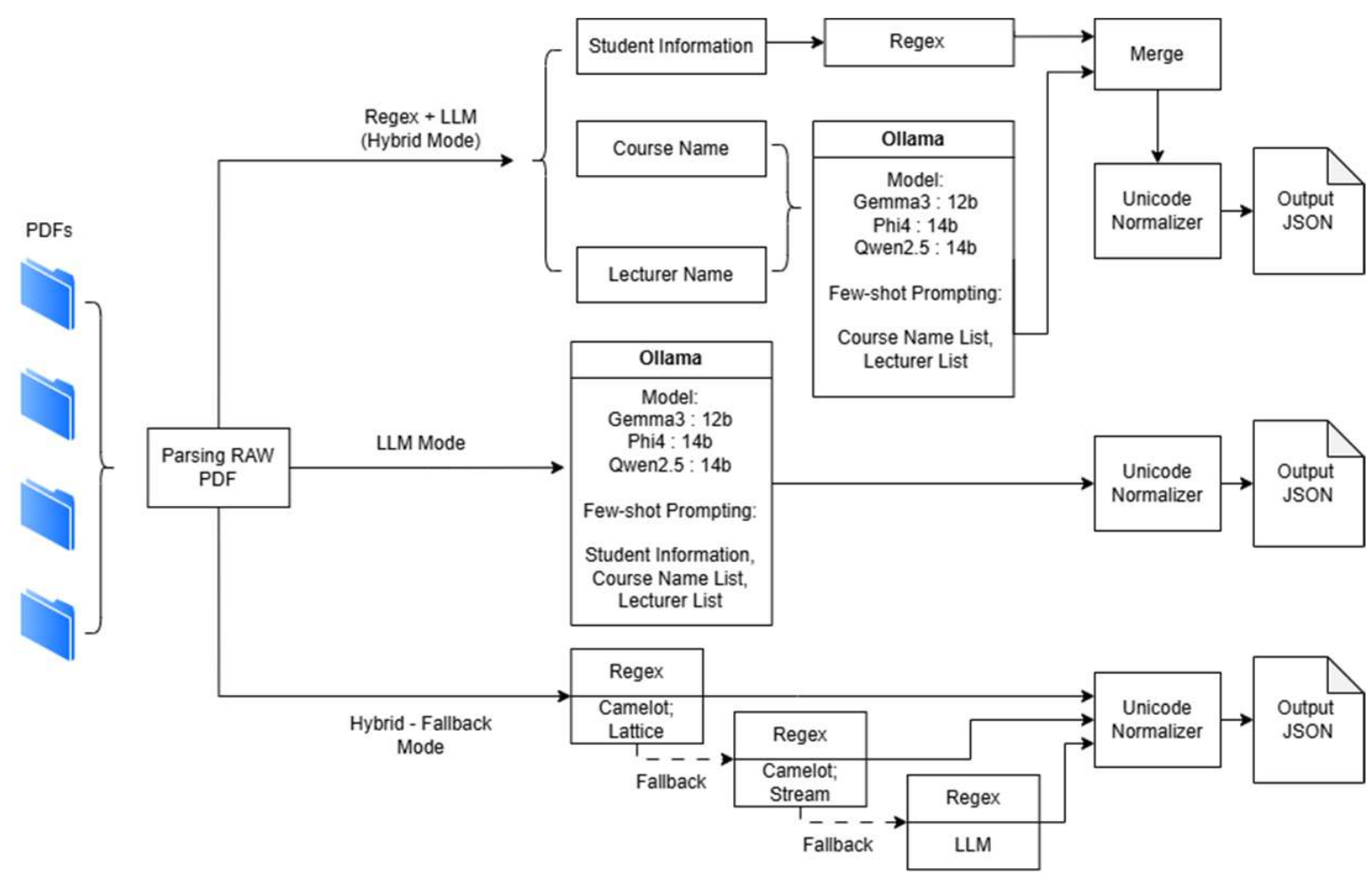

Figure 1. System Pipeline

```
You ONLY extract data that ACTUALLY EXISTS in the text.

ABSOLUTE RULES:
- DO NOT fabricate any data
- DO NOT add any courses
- Advisors are IGNORED
- If none → empty array
- Number of lecturers = Number of courses
- JSON IS VALID ONLY

FORMAT:
{{
"courses": [],
"lecturers": []
}}

e.g.

"courses": [
"High Voltage Engineering",
"Power Systems Practicum",
],
"lecturers": [
"Rindi Wulandari, S.ST., M.Si.",
"Taryo, ST., MT."
],

If there are repeated lecturer names, ALLOW THEM!
THE NUMBER OF LECTURERS MUST EQUAL THE NUMBER OF COURSES!
```

Figure 2. Few-shot Prompt

In designing the system pipeline, the primary consideration for the LLM was CTX = 3072, which estimated the number of tokens to be processed per KRS PDF page. TIMEOUT_HYBRID = 180 seconds, TIMEOUT_LLM = 300 seconds to accommodate potential computational requirements. In Figure 2, a few-shot prompt method was also used to direct LLM to produce output following explicit examples, with the expectation that the output would conform to the GT without any further complex settings. The following is a snippet of the prompt code used.

### 2.2. Evaluation Mechanism

The evaluation uses two main metrics: exact match (EM) and Levenshtein similarity. Levenshtein is used to tolerate minor errors (such as spaces or punctuation), with a threshold of 0.7 [12]. EM, on the other hand, provides clarity in comparisons with GT. Results are also analysed based on biographical data, course, and lecturer categories. The testing scheme is shown in the Figure 3.

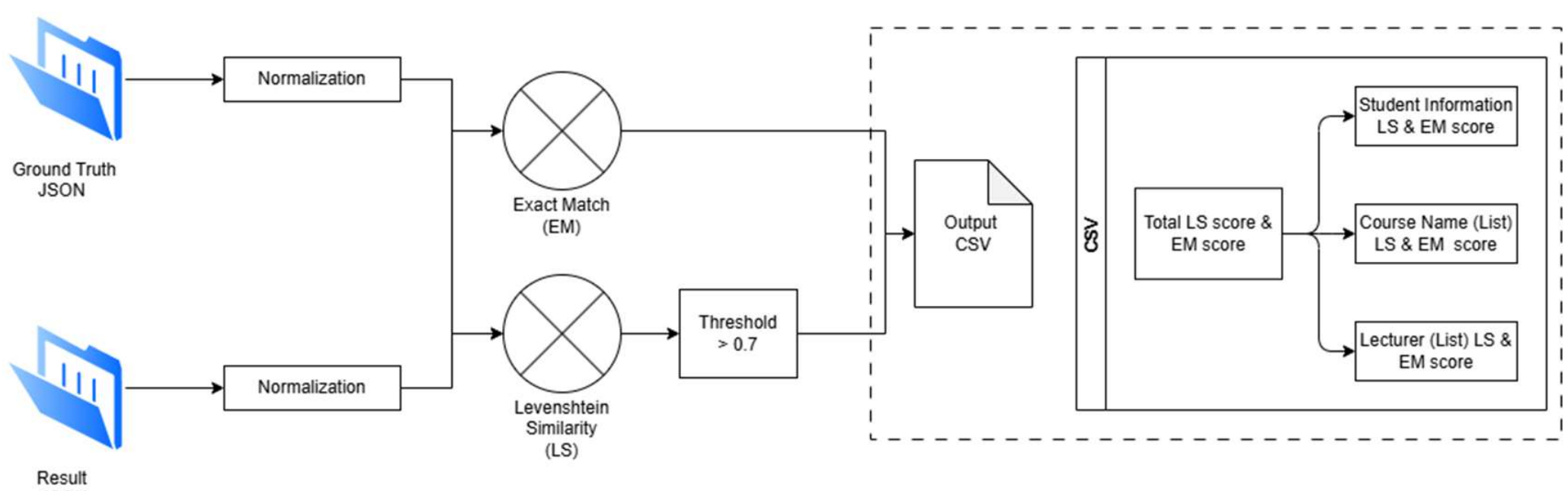

Figure 3. Evaluation Scheme

## 3. RESULT

The PDF parsing test results displayed in Figure 4 and Figure 5, compared to ground-truth (GT) JSON data, by study program category (35 PDFs per study program), show that: for the Electrical Engineering study program, exact match (EM) and Levenshtein Similarity (LS) scores were high in the LLM-only mode using the Phi4 and Qwen2.5 models. Conversely, the Gemma3 model showed particularly high scores in the hybrid mode (regex + LLM).

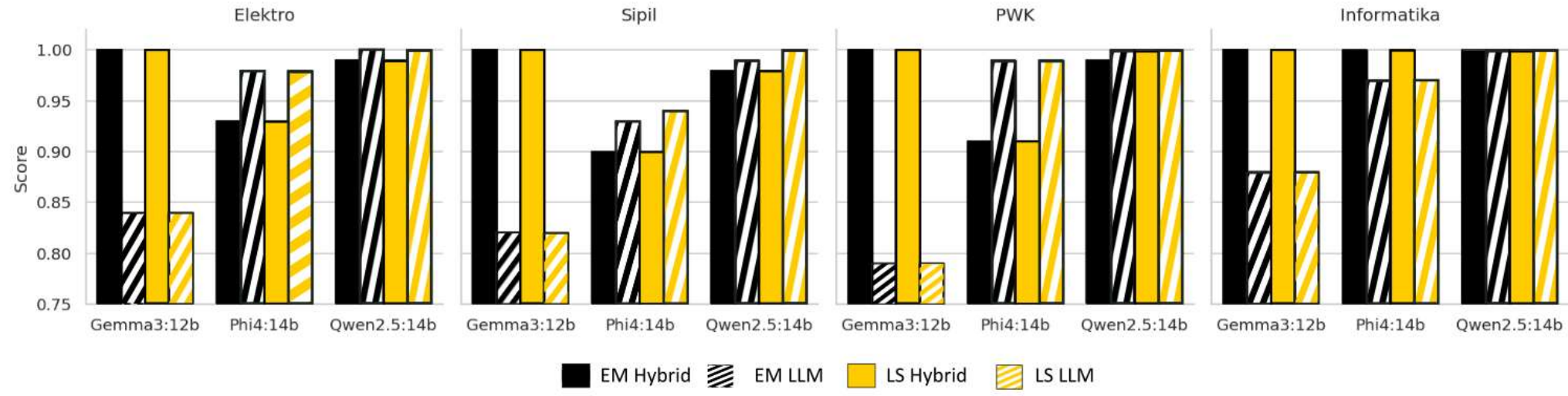

Figure 4. Exact Match (EM) and Levenshtein Similarity (LS) scores

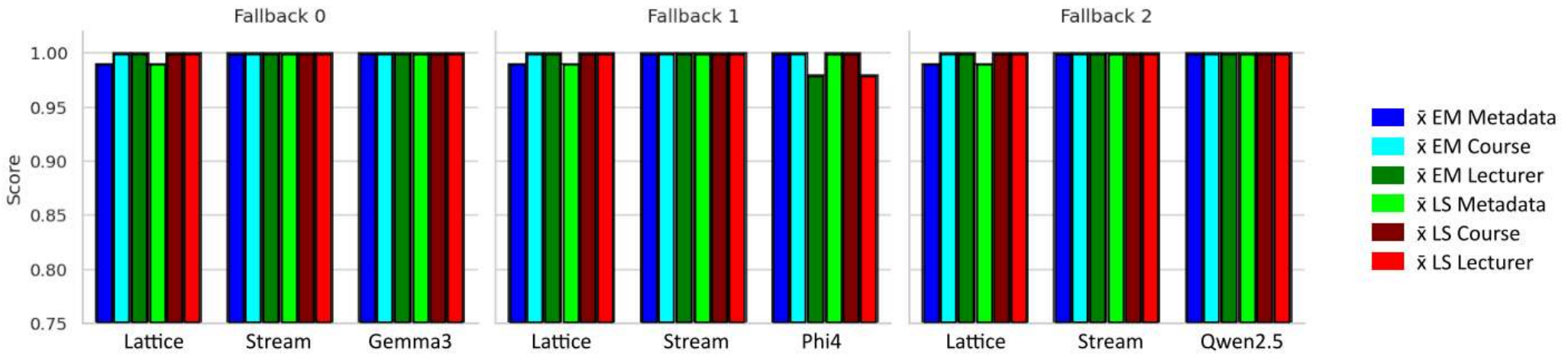

Figure 5. Fallback Exact Match (EM) and Levenshtein Similarity (LS) scores

In the Civil Engineering study program, Phi4 performed poorly compared to the other models in both the hybrid and LLM-only modes. Gemma3 performed very well in the hybrid mode but dropped significantly in the LLM-only mode. In the Urban and Regional Engineering study program, Phi4 performed well in the LLM-only mode, and Qwen2.5 scored the best across all modes. In the Informatics study program, the hybrid mode scored perfectly across all models, with Phi4 performing higher in the hybrid mode, and Qwen2.5 being the most accurate and stable.

Slight differences with GT, with a threshold of 0.7, are shown in the parsing scores for the Civil Engineering and Urban and Regional Planning (PWK) study programs, indicating a higher LS score than EM, specifically, in the LLM Phi4 and Qwen2.5 modes for the Civil Engineering study program, and in the hybrid mode in the Qwen2.5 model for the Urban and Regional Planning (PWK) study program.

In Table 1, the Qwen2.5 model demonstrates high accuracy and stability across all modes and study programs. The overall results for 140 sample PDFs also show scores per data category: metadata (student name, student ID, study program, semester, and academic year), the list of courses taken in that semester, and the list of lecturers. Table 1 shows parsing results that match the GT, with differences highlighted in red. Gemma3:12b cannot stand alone; in all study programs in LLM-only mode, all data categories have red indicators, but it is perfect in hybrid mode (with regex). For Phi4:14b, errors/differences with GT were minimal in the metadata and course lists, but significantly decreased in the lecturer list. The best result, with the fewest red flags and all scores above 0.9, was Qwen2.5:14 b. Qwen2.5 had the most errors in the hybrid mode course list.

Table 1. EM and LS scores

| Prodi | Mode | Model | $\bar{x}$ EM Metadata | $\bar{x}$ EM Course | $\bar{x}$ EM Lecturer | $\bar{x}$ LS Metadata | $\bar{x}$ LS Course | $\bar{x}$ LS Lecturer |
|---|---|---|---|---|---|---|---|---|
| Informatika | Hybrid | Gemma 3:12b | 1.000 | 1.000 | 1.000 | 1.000 | 1.000 | 1.000 |
| PWK | Hybrid | Gemma 3:12b | 1.000 | 1.000 | 1.000 | 1.000 | 1.000 | 1.000 |
| T. Elektro | Hybrid | Gemma 3:12b | 1.000 | 1.000 | 1.000 | 1.000 | 1.000 | 1.000 |
| T. Sipil | Hybrid | Gemma 3:12b | 1.000 | 1.000 | 1.000 | 1.000 | 1.000 | 1.000 |
| Informatika | LLM only | Gemma 3:12b | 0.829 | 0.909 | 0.909 | 0.829 | 0.909 | 0.909 |
| PWK | LLM only | Gemma 3:12b | 0.686 | 0.844 | 0.844 | 0.686 | 0.844 | 0.844 |
| T. Elektro | LLM only | Gemma 3:12b | 0.771 | 0.878 | 0.878 | 0.771 | 0.878 | 0.878 |
| T. Sipil | LLM only | Gemma 3:12b | 0.800 | 0.842 | 0.832 | 0.800 | 0.842 | 0.832 |
| Informatika | Hybrid | Phi 4:14b | 1.000 | 1.000 | 1.000 | 1.000 | 1.000 | 1.000 |
| PWK | Hybrid | Phi 4:14b | 1.000 | 1.000 | 0.756 | 1.000 | 1.000 | 0.756 |
| T. Elektro | Hybrid | Phi 4:14b | 1.000 | 0.977 | 0.819 | 1.000 | 0.977 | 0.823 |
| T. Sipil | Hybrid | Phi 4:14b | 1.000 | 0.996 | 0.722 | 1.000 | 0.996 | 0.722 |
| Informatika | LLM only | Phi 4:14b | 1.000 | 1.000 | 0.933 | 1.000 | 1.000 | 0.933 |
| PWK | LLM only | Phi 4:14b | 1.000 | 1.000 | 0.977 | 1.000 | 1.000 | 0.977 |
| T. Elektro | LLM only | Phi 4:14b | 1.000 | 1.000 | 0.958 | 1.000 | 1.000 | 0.958 |
| T. Sipil | LLM only | Phi 4:14b | 0.994 | 1.000 | 0.823 | 1.000 | 1.000 | 0.823 |
| Informatika | Hybrid | Qwen 2.5:14b | 1.000 | 1.000 | 1.000 | 1.000 | 1.000 | 1.000 |
| PWK | Hybrid | Qwen 2.5:14b | 1.000 | 0.996 | 1.000 | 1.000 | 0.999 | 1.000 |
| T. Elektro | Hybrid | Qwen 2.5:14b | 1.000 | 0.992 | 0.992 | 1.000 | 0.992 | 0.992 |
| T. Sipil | Hybrid | Qwen 2.5:14b | 1.000 | 0.983 | 0.983 | 1.000 | 0.983 | 0.983 |
| Informatika | LLM only | Qwen 2.5:14b | 1.000 | 1.000 | 1.000 | 1.000 | 1.000 | 1.000 |
| PWK | LLM only | Qwen 2.5:14b | 1.000 | 1.000 | 1.000 | 1.000 | 1.000 | 1.000 |
| T. Elektro | LLM only | Qwen 2.5:14b | 1.000 | 1.000 | 1.000 | 1.000 | 1.000 | 1.000 |
| T. Sipil | LLM only | Qwen 2.5:14b | 0.994 | 1.000 | 1.000 | 1.000 | 1.000 | 1.000 |

Table 2. Throughput

| Mode | Model | $\bar{x}$ Throughput (sec/PDF) | PDF/min |
|---|---|---|---|
| Hybrid | Gemma3 | 95.45 | 0.63 |
| Hybrid | Phi4 | 87.48 | 0.69 |
| Hybrid | Qwen2.5 | 93.55 | 0.64 |
| LLM Only | Gemma3 | 111.93 | 0.54 |
| LLM Only | Phi4 | 102.11 | 0.59 |
| LLM Only | Qwen2.5 | 108.08 | 0.56 |

Tests of a total of 140 PDF files (35 per study program) across two modes (hybrid, LLM-only) and three models (Gemma3, Phi4, Qwen2.5) required nearly 24 hours of computing time on a consumer-grade laptop (CPU-only). The throughput from the hybrid mode configuration (regex + LLM) showed that the computational load per PDF ranged from 95 to 108 seconds, or 1.5 to 1.8 minutes. Looking solely at throughput, without considering accuracy or stability, the highest was Phi4's hybrid mode (with regex), at 0.69 PDFs/minute, and the lowest was Gemma3's LLM-only mode. Meanwhile, in Table 2, the performance of Qwen2.5 is in the middle, namely 0.56 – 0.64 PDF/minute.

Table 3. Fallback Throughput

| Fallback Mode | Mode | $\bar{x}$ EM | $\bar{x}$ LS | n Data | $\bar{x}$ Throughput (sec) | $\bar{x}$ Throughput (with 1st attempt) |
|---|---|---|---|---|---|---|
| 0 | Lattice | 0.99 | 0.99 | 799 | 0.6338 | |
| 0 | Stream | 1 | 1 | 52 | 0.7567 | |
| 0 | Gemma | 1 | 1 | 9 | 28.1962 | 30.81 |
| 1 | Lattice | 0.99 | 0.99 | 799 | 0.6412 | |
| 1 | Stream | 1 | 1 | 52 | 0.7606 | |
| 1 | Phi | 0.99 | 0.99 | 9 | 29.7563 | 33.88 |
| 2 | Lattice | 0.99 | 0.99 | 799 | 0.6364 | |
| 2 | Stream | 1 | 1 | 52 | 0.759 | |
| 2 | Qwen | 1 | 1 | 9 | 32.1187 | 36.36 |

In addition to simple deterministic methods like regex and hybrid approaches with LLMs, the PDF parsing process was also tested with a non-LLM Python library that claims to be layout-aware, especially for table data (but limited to text-based PDFs, not images or OCR results), namely Camelot. This was done to evaluate the performance, effectiveness, and efficiency of various methods for parsing text-based PDFs. Camelot itself has several modes, or, as the official documentation calls them, "flavors". In testing, the lattice flavour was used because it aligns with the official documentation's recommendations: for tables with bold/clear line borders, use the lattice flavour; for tables without borders, use the "stream" flavour. These two flavours were chosen as fallbacks. Although there are other flavours, such as hybrid, the focus is on two flavours first to avoid widening the research scope while observing the test results.

Although claimed to be capable of parsing with layout-aware parsing, a second fallback was added to the pipeline, namely an LLM. Regex is still involved in parsing metadata, and Camelot is assigned to the data categories of course lists and lecturer names. In Table 3, the test showed a striking difference in computation and throughput, with results under 1 second per PDF. Therefore, the test was expanded to 860 KRS PDF files to obtain more comprehensive results. This was not done for the regex-LLM and LLM-only tests due to heavy computation and considerable time. Based on the previous test, 1 PDF took 1.5 minutes, so 860 PDFs would have taken about 21.5 hours, not to mention the various modes and model tests. In addition to the sharp difference in throughput, the average PDF parsing score relative to GT showed very high EM and LS, namely 0.99-1.00. Although Camelot successfully handled most of the parsing (as seen in n Data, 799 + 52 PDFs), the LLM fallback completed the remaining parsing, as 9 PDFs were handled by the LLM, indicating that the flavour lattice and stream were invalid in some cases. Interestingly, the LLM computation time in this pipeline is also shorter than in the previous modes (regex+LLM and LLM only), ranging from 28 to 33 seconds (excluding the first-attempt LLM, which typically requires a longer startup time). However, including the first attempt also ranges from 30 to 37 seconds. In this pipeline, Gemma3 and Qwen2.5 performed the highest, while Phi4 fell short on the LLM fallback. Flavour Lattice handled the most data, with an accuracy score of 0.99, while stream handled much less data, achieving a perfect score of 1.

A more detailed look at the parsing results by data category reveals failures in fallbacks 0 and 2, while the overall results for regex, Camelot Stream, and Gemma3 achieved perfect scores. However, there was a minor failure in metadata reading (score 0.99). For fallback 1, the Phi4 model failed to parse the metadata and the lecturer list. These results are still promising considering the total data is quite large, namely 860 PDFs and with a relatively short throughput for a consumer-class laptop, CPU only.

## 4. DISCUSSIONS

A small experiment using regex-only methods was conducted. Despite its high computational efficiency, it was unable to handle the variety and complexity of the data, resulting in very poor accuracy and difficulty generalising (e.g., across the variety of course codes or lecturer titles). Regex data is incapable of handling multi-line data and the occasional odd Unicode occurrences. Therefore, a specific test using regex alone was deemed unnecessary. Regex data was used for specific metadata groups. The rest was submitted or combined with other methods, which also helped reduce the computational burden.

To determine the model size, a small-sample test was initially conducted to identify a reasonable size for use on a CPU-only laptop, ranging from 1b to 7b. However, these tests yielded very poor results, so it was decided not to proceed, despite attempts to tune the prompt and adjust the context length and timeout. The model failed, for example, when parsing rows of data with varying course codes across study programs; it appeared confused about sorting the data or struggled with the layout. When the selection was increased to 12b–14b, the results were satisfactory and the best compromise for further testing. For fair testing, we would have preferred to compare models of the same size, but unfortunately, the 14b size was not available on the Ollama portal for Gemma3, so the closest size, 12b, was used instead. Ollama's use was intended to simplify the experiment without complex configurations.

The 0.7 threshold was determined based on reference considerations and sensitivity analysis of the results compared to GT. There were some tolerable differences with GT, such as the missing letter "a" in the middle of the word "perencanaan" (in the course category), or the missing letter "h" at the end of the lecturer's name, "Juariyah."

In testing the results, normalisation was performed because it would be too rigid to set (EM=1) due to the possibility of minor differences, such as spacing between the lecturer's name and their title, or the "." symbol in the title. This is because in some situations, for example, in the GT data, the original data is "Taryo, S.T, M.T", the parsed result "Taryo, S.T., M.T." contains a period at the end. This is considered tolerable because it does not damage the essence of the data, which is the lecturer's name, not the title.

In testing, the list of courses and their lecturers is limited to 15. A simple rule was created: the comparison result is scored 1 for an exact match (EM) if: GT exists and = parsed result; if GT is empty and the parsed result is also empty. Levenshtein Similarity (LS) is similar to EM and is calculated using the same formula, but with a tolerance limit of 0.7 to make the comparison results more realistic and less rigid.

Gemma3's score in the Electrical Engineering study program dropped when either the hybrid or LLM mode failed to read the data, leaving the JSON file empty. This automatically resulted in a 0 score. This failure was consistent across all data categories, with Gemma3 experiencing it only in LLM mode.

Qwen2.5 sometimes failed to capture multi-line text. For example, in the case of the PWK study program, the course title was long: "Analisis Sumber Daya dan Geologi Lingkungan," which caused the word "Lingkungan" to scroll down in the course PDF. Qwen2.5's hybrid parsing missed this word, resulting in a lower score (0.737). Qwen2.5 also missed one letter in the PWK study program's hybrid-mode courses: "sisteminformasiperencanaanlanjutan" vs "sisteminformasiperencananlanjutan" (LS score = 0.9706).

Phi4 in the Civil Engineering study program (LLM-only mode) also missed one letter in a student's name. For example, after normalisation, the GT was "dyniashalihatinalfitri," while the parsed result was "dynashalihatinalfitri" (LS score = 0.9545). In the Electrical Engineering study program, Phi4 also failed to read one letter in a lecturer's name: "jujujuhaeriyahsstmt" vs. "jujujuhaeriyasstmt" (LS score = 0.9474).

For Phi4, judging from the table, red indicators frequently appear in the lecturer name list. Judging from the parsing results, Phi4 omits the lecturer's name when the name read is also the advisor's name. The prompt does indeed instruct the student to ignore the advisor's name (meaning the one at the top of the table, as its data is not used), but Phi4 apparently removes it entirely. This condition occurs unevenly, as sometimes it is correct, sometimes it is deleted. This character does not occur in other models. Because a lecturer's name was removed, the order of subsequent lecturers on the list was affected, and the pairing of courses and their lecturers was also confused. This caused the Phi4 score to fall below Qwen2.5. The prompt also emphasised that the number of courses must equal the number of lecturers. This is difficult to control in Phi4.

Relying on LLMs' performance on a CPU-only laptop yields less-than-satisfactory throughput, as test results show it takes 1.5-1.8 minutes per PDF. This can be problematic when processing a large number of PDFs. However, the results are quite promising across EM and LS scores, computational load, and throughput in the third fallback mode, which relies on Camelot in addition to regex. LLM is used as a last resort when other methods fail to read/understand the data.

Failures occur when Lattice mode fails to read multi-line data, such as in the Civil Engineering study program and across other study programs. For example, the relatively long student names "mochamadnabielhaaritsfadillahh" vs "mochamadnabielhaarits" (LS score = 0.7). For failures in reading long names, if the LS score is <0.7, the value is considered zero. This occurred with several student names that were around 28 characters long after normalisation. Additionally, Phi4's LLM fallback discards duplicate lecturer names, resulting in a score of 0 (even though the prompt rule allows them). Because some lecturers teach more than one course, their names appear on the KRS multiple times, but Phi4 fails to recognise this.

The default mode for this fallback system is Lattice. Testing shows that the PDFs that show a small number of courses, typically 1-2, fall into the Stream fallback. This likely confuses Lattice when reading the data because the vertical borders are too close together (because the course list is short). The reason for the failure in Stream mode and the LLM fallback is unclear. However, upon visual inspection, the files that fall into the LLM fallback category have a small number of courses (typically 1-2) and a larger course column size than the other files, which likely impacts the parsing process. Test results show that LLM (Gemma3 and Qwen2.5) were able to read the data when Camelot failed. However, for Phi4, there was one failure due to a recurrence of a character that deleted a duplicate lecturer name, resulting in an incomplete total score.

Regex proved effective for stable biodata, but was not scalable for course tables due to high variation between study programs. LLM facilitates generalisation but requires deterministic control to maintain reliability.

Levenshtein similarity provides a more realistic metric than an exact match, though it requires a threshold to avoid bias. Some observed failures include missing table rows and incorrect course-lecturer pairs.

## 5. CONCLUSION AND FUTURE WORK

Conclusion. This study demonstrates that a purely LLM-based information extraction approach has strong generalisation capabilities but requires substantial computational resources and exhibits reliability differences across models. Integrating deterministic methods (e.g., regex) with an LLM in a hybrid scheme has been shown to improve the stability of results, particularly for metadata elements with relatively fixed patterns. However, CPU-only throughput is relatively low, averaging 1.5–1.8 minutes per PDF.

Further experiments utilising a layout-aware library (Camelot) for parsing text-based tables demonstrated significant efficiency improvements. Most documents could be processed with the lattice flavour at near-perfect accuracy, while stream and LLM fallbacks were required only in a small subset of cases. The progressive fallback approach (Lattice → Stream → LLM) achieved the optimal balance between accuracy, stability, and computational efficiency.

Overall, the Qwen 2.5:14b model demonstrated the most stable performance across various configurations. These results indicate that the combination of deterministic methods, layout-aware parsing, and neural fallback is a more robust strategy than a single LLM-based approach for tabular PDF information extraction, as demonstrated on the KRS case study.

Future Work. Several further development directions are possible to improve the system's scalability and generalisation. First, experiments in a GPU-based environment are needed to evaluate throughput improvements and energy efficiency. With GPU acceleration, LLM inference time is expected to be significantly reduced, allowing evaluation on larger datasets without compromising computational time. Comparing CPU-only versus GPU-enabled inference is also important for assessing the cost-performance trade-off. Second, further research could explore model optimisation through quantisation, parameter-efficient fine-tuning, or the use of lighter architectures that remain robust for understanding complex table structures. A domain-specific fine-tuning approach to Indonesian academic documents also has the potential to improve reliability across multiple rows and across varying academic degrees. Third, expanding the study to image-based PDF documents (scanned/OCR) would broaden the system's application scope. Robust OCR integration and evaluation on cross-institutional documents could test the pipeline's generalizability. With this development, the hybrid deterministic–neural information extraction approach has the potential to be an efficient, private, and scalable solution for tabular PDF document processing across a range of application domains.